\documentclass{article}

\usepackage[colorlinks,bookmarksopen,bookmarksnumbered,citecolor=red,urlcolor=red]{hyperref}
\usepackage{tikz}
\usepackage{jfrc}
\usepackage{amsmath,amssymb}
\usepackage{MnSymbol,enumitem}

\newtheorem{example}{Example}
\newtheorem{remark}{Remark}
\newtheorem{definition}{Definition}

\usepackage{authblk}

\title{Optimal Algorithm Allocation for Single Robot Cloud Systems\thanks{© 2021 IEEE.  Personal use of this material is permitted.  Permission from IEEE must be obtained for all other uses, in any current or future media, including reprinting/republishing this material for advertising or promotional purposes, creating new collective works, for resale or redistribution to servers or lists, or reuse of any copyrighted component of this work in other works.}}
\author[1]{Saeid Alirezazadeh}
\author[2]{Lu\'{i}s A.~Alexandre}
\affil[1]{C4-Cloud Computing Competence Center, Universidade da Beira Interior, C4-Estrada Municipal, 506, 6200-284, Covilh\~{a}, Portugal.}
\affil[2]{NOVA LINCS, Universidade da Beira Interior, Covilh\~{a}, Portugal.}
\affil[1]{saeid.alirezazadeh@gmail.com}
\affil[2]{lfbaa@di.ubi.pt}
\date{}

\providecommand{\keywords}[1]{\textbf{\textit{Keywords---}} #1}
\begin{document}

\maketitle

\begin{abstract}
For a robot to perform a task, several algorithms must be executed, sometimes simultaneously. The algorithms can be executed either on the robot itself or, if desired, on a cloud infrastructure. The term cloud infrastructure refers to hardware, storage, abstracted resources, and network resources associated with cloud computing. Depending on the decision of where the algorithms are executed, the overall execution time and memory required for the robot, change accordingly. The price of a robot depends on its storage capacity and computational power, among other factors. We answer the question of how to maintain a given performance and deploy a cheaper robot (lower resources) by allocating computational tasks to the cloud infrastructure depending on memory, computational power, and communication constraints. Even for a fixed robot, our model provides a way to achieve optimal overall performance. We provide a general model for optimal algorithm allocation decision under certain constraints. We illustrate the model with simulation results. The main advantage of our model is that it provides optimal task allocation simultaneously for memory and time.
\end{abstract}

\keywords{Cloud robotics, cloud computing, fog computing, edge computing, memory and time costs, optimization strategy, allocation algorithm.}

%

\section{Introduction}
recent decades, various aspects of human life have been greatly enhanced by the use of robotic systems, such as robots for industrial and manufacturing purposes \cite{ifr:2020}, military robots \cite{nath:2014}, domestic robots \cite{xu:2014}, and others \cite{bruno:2016}. Robots have some level of intelligence to perform various tasks automatically. Cloud robotics is described as a way to overcome some of the computational limitations of robots by using the Internet and cloud infrastructure to delegate computations and also share large data in real time \cite{kehoe:2015}. An important factor in determining the performance of cloud-based robotic systems is deciding whether a task should be uploaded to the cloud, processed on a server (fog computing \cite{bonomi:2012}), or executed on the robot (edge computing \cite{shi:2016}), the so-called allocation problem. Our goal is to provide an answer to the allocation problem for a cloud robotic system with a single robot.

Saha and Dasgupta \cite{saha:2018} examined published works on cloud robotics from 2012 to 2018 and found that task assignment is still a challenge. The works \cite{wang:2014} 
 and \cite{kong:2017} are mainly concerned with optimizing the cost for robots to fetch all the necessary resources, assuming that the resources (algorithms and data) are allocated in advance. None of these studies answer the question of how to allocate resources optimally, but they have tried to find a way to reduce the cost when the resource allocation is fixed in advance.

The paper is organized as follows: In Section 2, we briefly describe related works on the allocation problem. In Section 3, we describe the graph theory notions used throughout the paper. In Section 4, we translate the time and memory optimizations as optimizations using time and memory algebras. Then, to find a joint optimization of memory and time, the problem is translated into finding a point with the minimum distance to the origin in a two-dimensional space. In Section 5, we present the algorithm for optimizing memory and time. In Section 6, we use simulation data to test our method and compare it with \cite{li:2018}, and in Section 7, we draw a conclusion.

\section{Related Works}
In a multi-robot system, the set of all tasks that can be performed by the system, $T$, is finite. At each instant, the system performs a set of tasks, $T_1$, which is a subset of the set of all tasks. At the same time, a new set of tasks, $T_2$, arrives that must be performed by the system. As time progresses, this process repeats. As can be seen in Figure~\ref{fig0}, there are two ways to assign tasks:
\begin{figure*}[t]\centering
\begin{tikzpicture}
\begin{scope}[every node/.style={rectangle,thick,draw}]
    \node (A) at (-1,0) [text width=7cm] {Task allocation problem: \\$T=\{A_1,\ldots,A_m\}$ and  $(T_i)_{i\in\mathbb{N}}=T_1,T_2,\ldots, \subseteq T$};
    \node (B) at (-4.5,-2) [text width=5cm] {Dynamic: Optimal performance for allocating $(T_i)_{i\in\mathbb{N}}$};
    \node (C) at (1,-2) [text width=3cm] {Static: Optimal performance for allocating $T$};
    \node (D) at (-6,-4) [text width=3cm] {Centralized: Central unit provides task allocation, \cite{Burkard:2012, Gombolay:2018}};
    \node (E) at (-2.6,-4.5) [text width=3cm] {Distributed: tasks disperse to all robots, and robots decide whether to perform tasks or not};
    \node (F) at (2.5,-6) [text width=2.5cm] {Combinatorial optimization-based, \cite{Gombolay:2013, Wang:2015}} ;
    \node (G) at (-0.5,-6.5) [text width=2.4cm] {Behavior-based, \cite{Parker:1998, Chen:2018}} ;
    \node (H) at (-3.2,-6.5) [text width=2.2cm] {Market-based, \cite{Alaa:2013, Wang:2017, wang:2014}} ;
    \node (I) at (-6.2,-6) [text width=2.5cm] {Evolutionary algorithm-based, \cite{Lane:2018, Arif:2017, Cheng:2015}} ;
    \node (J) at (0.5,-4) [text width=2.5cm] {Multi-robot, time optimization \cite{li:2018}};
    \node (K) at (4,-4) [text width=3.15cm] {Single robot, simultaneously time and memory optimization (Our result.)};
\end{scope}

    \path [->] (A) edge[thick,->] (B);
    \path [->](A) edge [thick,->] (C);
    \path [->](B) edge [thick,->] (D);
    \path [->](B) edge [thick,->] (E);
    \path [->](E) edge [thick,->] (F);
    \path [->](E) edge [thick,->] (G);
    \path [->](E) edge [thick,->] (H);
    \path [->](E) edge [thick,->] (I);
    \path [->](C) edge [thick,->] (J);
    \path [->](C) edge [thick,->] (K);
    \path [->](J) edge [thick, dashed,->] (K);
\end{tikzpicture}
\caption{Diagram of the studies on task allocation problem. The dashed arrow is used only to represent that, as we explain later in this paper, our result for a single robot is not a specific instance of the result of \cite{li:2018}.}
\label{fig0}
\end{figure*}
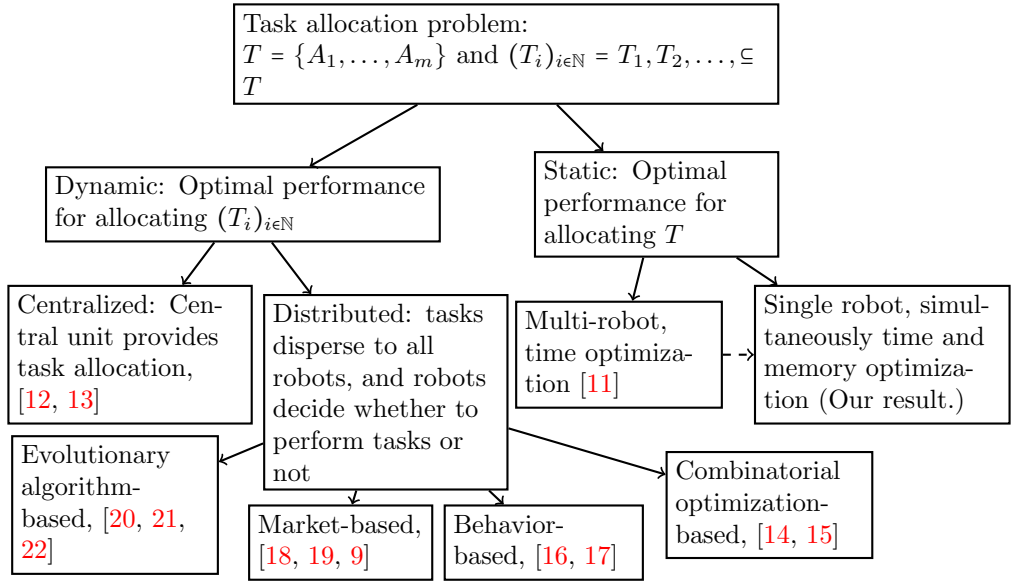
\begin{itemize}
\item a \textbf{dynamic task allocation} answers the question of how to achieve the optimal performance of the system by dynamically allocating tasks according to time, $(T_i)_{i\in\mathbb{N}}$, in the order of the sets of arrived tasks.
\item a \textbf{static task allocation} answers the question of how to achieve the optimal performance of the system by allocating the tasks in the set of all tasks, $T$.
\end{itemize}
A static task allocation provides the minimum cost of the system after all tasks have been performed, while a dynamic task allocation provides the minimum cost of the system in the time horizon. Moreover, for the same set of tasks, the minimum cost obtained by static task allocation is the largest lower bound of the minimum cost obtained by dynamic task allocation.

Dynamic task allocation is classified as a centralized or a distributed assignment. In centralized methods, a central planning unit that has information about the entire environment handles task allocation, \cite{Burkard:2012}. In the work in \cite{Gombolay:2018}, a centralized algorithm called ''Tercio'' was presented for task allocation that minimizes latency and physical proximity to tasks. In distributed assignment, instead of a centralized unit, all tasks are distributed to all robots, and the robots decide which tasks to perform. Four main approaches were used for distributed assignment: 
\begin{itemize}
\item behavior-based: based on the problem features, it is decided whether the robot should consider a task or not, \cite{Parker:1998};
\item market-based: based on an auction-based mechanism, \cite{Alaa:2013};
\item combinatorial optimization-based: transforms the problems into combinatorial optimization problems and uses a suitable existing technique to solve them, \cite{Gombolay:2013};
\item evolutionary algorithm-based: uses evolutionary operators on the space of solutions, \cite{Lane:2018, Arif:2017}.
\end{itemize}
For task allocation in a cloud robotic system, most studies only move computationally intensive tasks to the cloud without considering the communication between robots and the cloud infrastructure, \cite{davinci:2010, Slam:2015, Rapyuta:2013}.

For dynamic task allocation, \cite{Maimon:1986} proposed a dynamic task allocation in which a task is assigned to a robot based on the availability and suitability for the task and the future plan; \cite{Chen:2018} proposed an algorithm to optimize the latency, energy consumption, and computational cost considering the characteristics of cloud robotic architecture, task characteristics, and quality of service issues for cloud robotics; \cite{Cheng:2015} proposed an architecture for wearable computing devices and investigated offloading strategies that preserve the required delays based on the characteristics of the devices, and proposed a fast algorithm based on the genetic algorithm as an offloading strategy; \cite{Wang:2015} studied the resource sharing problem in a cloud robotic system and presented a resource management strategy with near real-time data retrieval; \cite{Wang:2017, wang:2014} studied real-time resource allocation based on a hierarchical auction-based method; \cite{Gerkey:2004} studied dynamic task allocation, and they classified robots, tasks, and methods of task assignment to robots; \cite{Korsah:2013} extended the work of \cite{Gerkey:2004} for dynamic task scheduling and studied the different degree of interrelatedness dependency between arrived tasks and tasks scheduled in a robot to be allocated to and the other tasks scheduled in other robots, and for each degree they proposed a general optimization model that can be used to solve the allocation problem; \cite{Schillinger:2018} proposed a method for centralized dynamic task allocation based on modeling the problem as a non-deterministic finite state automaton, and a minimum cost solution is presented that depends on the energy level of the robots; \cite{Cai:2000} proposed a method for dynamic task allocation in which the allocation of tasks to a processor is done under the assumption that some tasks require multiple processors to be performed; \cite{Chopra:2017} studied dynamic task allocation and provided an extension of Kuhn's algorithm, see \cite{Burkard:2012}, by considering a distributed version of the method that allows a team of robots to cooperatively compute the optimal solution to a linear objective function without a coordinator or shared memory; \cite{Zhang:2018} proposed a method to handle the deadline of a task and also minimize the total cost; \cite{Chen:2019} studied task allocation under the assumption that the number of robots changes and the tasks are modifiable; and \cite{ WANG:2020} studied the optimal task allocation for the case where two collaborative teams of robots jointly perform some complex tasks. They provide the optimization statement formulation of their model based on set-theoretic parameters.

Static task allocation can be considered as the primary evaluation of a cloud robotic system since a solution obtained from static task allocation provides an optimal way to distribute the necessary algorithms for the execution of all tasks among the nodes of the cloud robotic system in the sense that all the necessary inputs for the execution of a task are available in the fastest way for the node to which the task is assigned. Therefore, it ensures that a node to which a task is assigned gathers all the required information in an optimal manner. In other words, static task allocation is not only as important as dynamic task allocation, but it also shows how to achieve the optimal performance of cloud robotic systems while performing any task. For static task allocation, the work in \cite{Lin:1995} studied static algorithm allocation (in a multi-robot system). However, cloud infrastructure and communication times are not considered in their study. As far as we can tell, the only work that addressed a similar problem is \cite{li:2018}. They used a similar technique to ours to solve the allocation problem. These authors only considered minimizing the overall execution time but without fully considering the communication. Moreover, they considered only the single parameter of time, but in our method, we also consider the parameter of memory usage of the robot. The approaches also differ in terms of the start and end of time: since all algorithms produce results that are needed by the robots, the time should range from when a robot sends a request to execute an algorithm to when it receives the result of that algorithm. In this way, the start and end points of the timing are the robot (if there are multiple robots, all robots are start and end points), but in \cite{li:2018}, the start and end points of timing for each algorithm depend on where they are executed. We will compare the two methods using detailed examples. In addition, the memory limitation of robots has a direct impact on latency, which is not considered in \cite{li:2018}. For example, if we replace the robot with a similar robot that has a smaller memory, the new robot may not be able to execute some of the required algorithms.

We also answer the question of optimized memory usage of the robot. Our methods for finding an optimized allocation of algorithms yield two classes each for memory and time at a given value, with the intersection yielding the optimal solution to minimize the robot's Memory-Time problem. 

\begin{figure*}[t]\centering
\begin{tikzpicture}
\begin{scope}[every node/.style={rectangle,thick,draw}]
    \node (A) at (0.5,1) [text width=14cm] {Necessary tasks to be performed on the Edge (E), the Fog (F), and the Cloud (C)\\ 1) Algorithm (1): Complexity $O_j(A_1)$, Response Time $T_j(A_1)$, Space necessary $S_j(A_1)$, ... ;\\$\vdots$\\n) Algorithm (n): Complexity $O_j(A_n)$, Response Time $T_j(A_n)$, Space necessary $S_j(A_n)$, ... .\\ Constraints on $O_j(A_i)$, $T_j(A_i)$, and $S_j(A_i)$'s for $i=1,\ldots,n$ and $j=E, F, C$.};
    \node[rectangle,dashed,draw] (B) at (-4.3,-3.2) [text width=6.4cm] {Identify the three sets partition $\{E,F,C\}$ of the set $\{1,\ldots,n\}$ such that:\\- if $e\in E$, then the algorithm $A_e$ must be performed on the Edge;\\- if $f\in F$, then the algorithm $A_f$ must be performed on the Fog;\\- if $c\in C$, then the algorithm $A_c$ must be performed on the Cloud;\\};
    \node (C) at (0.5,-1.7) [text width=1.25cm,align=center] {Decision};
    \node (D) at (0.5,-2.5) [text width=0.7cm,align=center] {Edge};
    \node (E) at (3.4,-2.5) [text width=0.55cm,align=center] {Fog};
    \node (F) at (6.3,-2.5) [text width=0.85cm,align=center] {Cloud} ;
    \node (G) at (0.6,-3.2) [text width=0.85cm,align=center] {Robot} ;
    \node (H) at (3.4,-3.2) [text width=0.9cm,align=center] {Server} ;
    \node (I) at (0.6,-4.3) [text width=2.4cm] {Perform algorithms $A_e$'s for $e\in E$} ;
    \node (J) at (3.4,-4.3) [text width=2.4cm] {Perform algorithms $A_f$'s for $f\in F$};
    \node (K) at (6.3,-4.3) [text width=2.4cm] {Perform algorithms $A_c$'s for $c\in C$};
\end{scope}

    \path [->](A) edge [thick,->] (C);
    \path [<->](B) edge [thick,<->] (C);
    \path [<->](C) edge [thick,<->] (D);
    \path [<->](C) edge [thick,<->] (E);
    \path [<->](C) edge [thick,<->] (F);
    \path [->](D) edge [thick, dotted,-] (G);
    \path [->](E) edge [thick, dotted,-] (H);
    \path [->](G) edge [thick,->] (I);
    \path [->](H) edge [thick,->] (J);
    \path [->](F) edge [thick,->] (K);
\end{tikzpicture}
\caption{A single robot model. The decision, as the optimal solution for the distribution of all algorithms, is shown in the dashed box. The response time is the sum of communication time and execution times.}
\label{fig3}
\end{figure*}
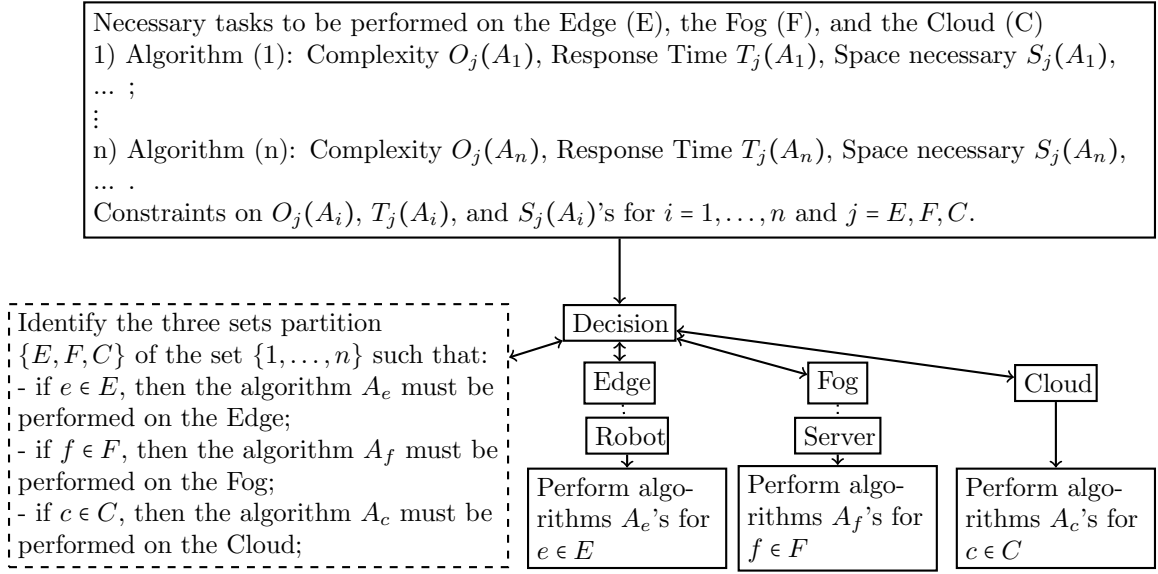

Figure~\ref{fig3} shows a general description of the problem we address. Throughout this paper, because of existing random delays, the communication times between the edge, fog, and cloud is considered as the average communication time, and the execution time of an algorithm on the edge, fog, and cloud is considered as the average execution time of the algorithm on the edge, fog, and cloud, respectively.

We start by constructing a general theoretical model, based on which we build the problem and find an optimal solution(s). Note that some of the algorithms can share information and some require the results of others. The model we describe here has similarities to the semi-lattice\footnote{A (semi-)lattice is a partially ordered set in which for any two elements there is a least upper bound (or a greatest lower bound).}, see \cite{birkhoff:1967}. This is achieved by including the elements $\mathbf{1}$ and $\mathbf{0}$ as the largest and smallest elements in the structure, respectively. See Figure~\ref{fig4}.
\section{Preliminaries}
Before describing the model and solution, we recall some mathematical concepts that we use throughout the manuscript to construct models.
\begin{definition}\label{def:def1}
We construct a directed graph $G=\left(V,\overrightarrow{E}\right)$ to help us formulate a general model. A directed graph $G=\left(V,\overrightarrow{E}\right)$ is defined by the set of vertices of the graph, which is the set of algorithms $V=\left\{A_1,\ldots,A_n\right\}$, and the set of directed edges, i. e., a subset of ordered pairs of elements of $V$, $\overrightarrow{E}=\left\{(A_i,A_j)\mid A_j \text{ uses the output of } A_i\right\}$. 
\end{definition}
\begin{definition}
Consider a directed graph $G=\left(V,\overrightarrow{E}\right)$ and $v\in V$. Then define: 
\begin{itemize}
\item the out-degree of $v$ is the number of elements of $\overrightarrow{E}$ in which $v$ occurred as the first component:
$$\mathrm{OutDegree}(v)=\left|\left\{w\in V\mid(v,w)\in\overrightarrow{E}\right\}\right|.$$
\item the in-degree of $v$ is the number of elements of $\overrightarrow{E}$ in which $v$ occurred as the second component:
$$\mathrm{InDegree}(v)=\left|\left\{w\in V\mid(w,v)\in\overrightarrow{E}\right\}\right|.$$
\end{itemize}

In a directed graph, two vertices $A_i$ and $A_j$ are called connected (or adjacent) if at least one of the edges $(A_i,A_j)$ or $(A_j,A_i)$ lies in $\overrightarrow{E}$.
\end{definition}
\begin{definition}
A subgraph of a graph $G$ is the graph obtained by removing some vertices of $G$, all directed edges to which the removed vertices are connected, and some other edges. 
\end{definition}
\begin{definition}
A cycle is a graph such that, starting from any vertex and following the direction of the edges, one can visit all vertices and edges once and return to the vertex from which one started\footnote{There is a slight difference between a cycle as a graph and a cycle in a graph. A cycle in a graph is a sequence of length greater than $3$ of adjacent vertices without repetition, except that the first and last vertices are the same.}.
\end{definition}
\begin{remark}\label{rem:rem1}
Some of the vertices of the directed graph in Definition~\ref{def:def1} must have in-degree $0$, and some others must have out-degree $0$. Also, the graph must not have a cycle as a subgraph. Otherwise, since all vertices are assumed to have non-zero in-degree, this means that since $V$ is a finite set, $G$ has a cycle subgraph. This means that there is a subset of algorithms whose execution cannot be started until we have the final result of their execution, which is a contradiction.
\end{remark}%
\section{Proposed Method}
Like other algorithms searching for the longest path in a graph, our method's time complexity is in NP.
\subsection{Time Optimization}
\begin{figure}[t]\centering
\includegraphics[width=0.6\linewidth]{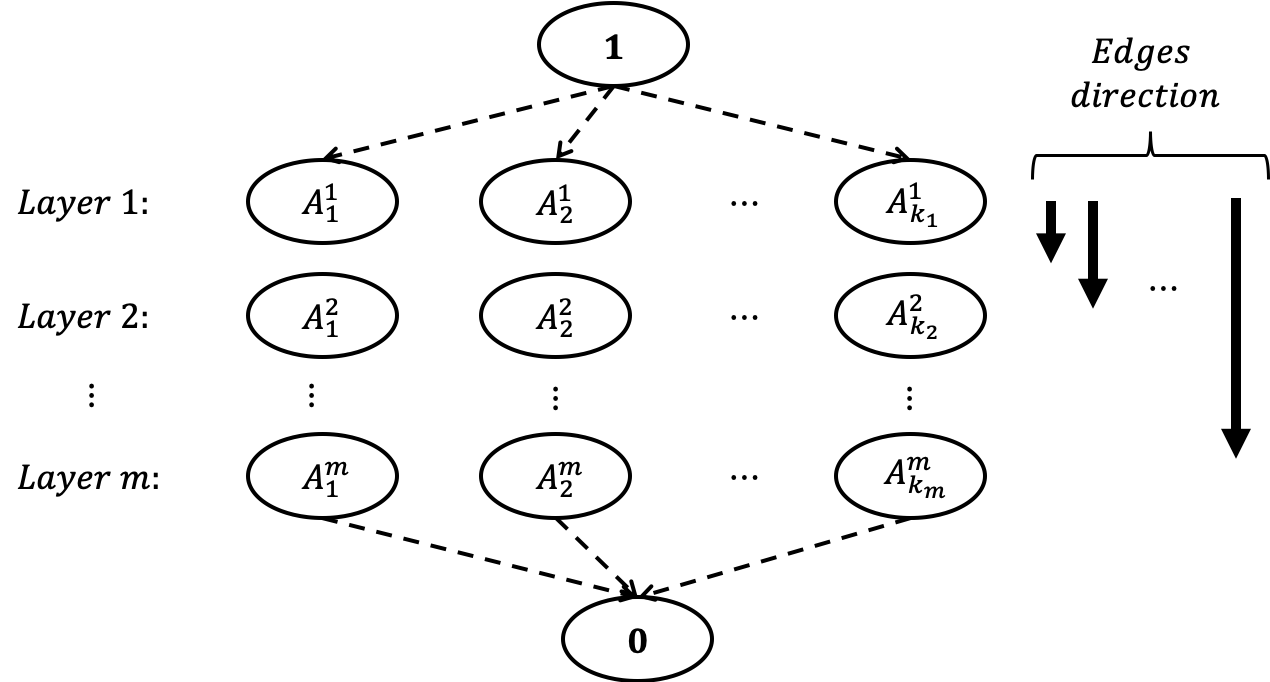}
\caption{Graph with algorithm execution dependence, shown with downward edges. Adding the virtual vertices $\mathbf{0}$ and $\mathbf{1}$ to the graph (we have assumed here that the graph itself is a connected graph), it becomes a semi-lattice. $A_i^j$ represents the algorithm $A_i$ at layer $j$. Note that the $k_i$ for $i=1,\ldots,m$ are not necessarily equal.}
\label{fig4}
\end{figure}

The response time of an algorithm is the sum of the following times:
\begin{itemize}
\item the time required for the algorithm to produce a result independently of other algorithms, assuming that the necessary conditions for the algorithm to run are satisfied;
\item the average time required to transmit the result of the algorithm to the location where the algorithm is requested;
\item the average time required to transmit the request for the algorithm to the location where the algorithm is allocated.
\end{itemize}
By Remark~\ref{rem:rem1}, we can represent the graph such that all edges are downward, which can be constructed as follows: The vertices of the graph appear in different layers, the first layer consists of all vertices with in-degree zero. The second layer consists of all vertices such that there are only edges from vertices of the first layer to them. The next layer consists of all vertices such that there are only edges from vertices in the previous layers to them. It is clear that the last layer is the set of vertices whose out-degrees are zero, see Figure~\ref{fig4}. The constructed graph with downward edges can be viewed as a union of its connected components\footnote{A connected component is the weakly connected component in a directed graph.}. By adding virtual vertices $\mathbf{0}$ and $\mathbf{1}$ to each of the connected components of the graph with vertex $\mathbf{1}$ at the top of the first layer and edges from it to all vertices in the first layer and vertex $\mathbf{0}$ at the bottom of the last layer and edges from all vertices in the last layer to it, this process turns the graph into a union of semi-lattices, denoted by $\mathcal{ SL }(G)$. We abuse the notation slightly and denote the virtual vertices of all connected components of the graph by $\mathbf{0}$ and $\mathbf{1}$.
\begin{remark}
The reason for using connected components instead of the whole graph is that the overall execution time of each connected component with respect to any allocation of algorithms is independent of the overall execution time of the other connected components, and having fewer vertices reduces computations.
\end{remark}

In Figure~\ref{fig4}, the algorithms in the nodes are different algorithms. That is, $A^x_y\neq A^j_i$ if at least one of the inequalities $j\neq x$ and $i\neq y$ holds.

Note that an execution flow from $\mathbf{1}$ to $\mathbf{0}$ in $\mathcal{ SL }(G)$ is a sequence of algorithms $1A'_1\cdots A'_k0$, where:
\begin{itemize}
\item elements $A'_i$'s for $i=1,\ldots,k$ are elements of $V$;
\item ordered pairs $(A'_i,A'_{i+1})$'s for $i=1,\ldots,k-1$ are elements of $\overrightarrow{E}$; 
\item the corresponding vertices $A'_1$ and $A'_k$ in the graph $G$ have zero in-degree and out-degree respectively;
\item since we consider a unique flow, for $i=1,\ldots,k$ the algorithm $A'_i$ will run in layer $i$ of $\mathcal{ SL }(G)$.
\end{itemize} 
\begin{remark}\label{rem:rem2}
If $(A_i, A_j)$ is a directed edge of $\mathcal{ SL }(G)$, then by construction, algorithm $A_j$ can only be executed after algorithm $A_i$ has been executed. Thus, the execution time of serial algorithms is the sum of the execution time of the individual algorithms.
\end{remark}

Denote by $\mathrm{ExecutionFlows}(G)$ the set of all execution flows from $\mathbf{1}$ to $\mathbf{0}$ in $\mathcal{ SL }(G)$.

For a given set of algorithms $\{A_1,\ldots,A_n\}$, generate the three $n$-tuples of the response times of the algorithms to be executed on the edge, the fog, or the cloud:
\begin{align*}
T_E=\left(T_E(A_1),\ldots,T_E(A_n)\right)&, T_F=\left(T_F(A_1),\ldots,T_F(A_n)\right),\\
T_C=\left(T_C(A_1),\ldots,T_C(A_n)\right)&.
\end{align*}
The mapping: 
\begin{equation}\label{eq:eq1}
\pi:\left\{A_1,\ldots,A_n\right\}\longrightarrow\left\{E,F,C\right\},
\end{equation}
is defined by 
\[\pi(A_i)=\left\{
\begin{array}{ll}
E&,\text{If the algorithm } A_i \text{ is going}\\
&~~\text{to be executed on the edge},\\
F&,\text{If the algorithm } A_i \text{ is going}\\
&~~\text{to be executed on the fog},\\
C&,\text{If the algorithm } A_i \text{ is going}\\
&~~~\text{to be executed on the cloud}.
\end{array}\right.
\]

Optimal solutions are obtained by finding suitable associated values for the mapping $\pi$. Then, for an algorithm $A_i$ and a known mapping $\pi$, $T^{\pi}(A_i)=T_{\pi(A_i)}(A_i)$ holds. Hence, $T^{\pi}(P_i)=\sum_{x=1}^kT_{\pi(A_x)}(A_x)$, and for all possible $\pi$, 
\begin{equation}\label{eq:eq2}
T(P_i)=\left\{T^{\pi}(P_i)\mid \forall\pi:\{A_1,\ldots,A_n\}\longrightarrow\{E,F,C\}\right\}.
\end{equation}

Equation \eqref{eq:eq2} means to find a set of all times for the execution of algorithms on $P_i$ by considering all possible ways of distributing algorithms for execution on the edge, the fog, or the cloud. Note that each element of $\mathrm{ExecutionFlows}(G)$ represents the relationship between algorithms. For a fixed mapping $\pi$, find the set of response times of execution flows with respect to the mapping $\pi$ from all possible mappings of \eqref{eq:eq1}:
\begin{equation}\label{eq:eq3}
\begin{array}{l}
\mathrm{ExecutionFlowsTimes}_{\pi}(G)=\\
~~~~\left\{T^{\pi}(P_i)\mid P_i\in\mathrm{ExecutionFlows}(G)\right\}.
\end{array}
\end{equation}
In the set \eqref{eq:eq3} with respect to the mapping $\pi$, we can use $\max\left\{\mathrm{ExecutionFlowsTimes}_{\pi}(G)\right\}$ to determine the maximum time needed for the robot to obtain the results of all algorithms. Note that for different mappings $\pi_1$ and $\pi_2$, the maximum may not occur for the same execution flow, even if we have multiple execution flows producing the maximum value. Therefore, the problem of distributing algorithms among the edge, the fog, and the cloud such that the minimum total time is consumed for the execution of all algorithms can be found by 
$\min\left\{\max\left\{\mathrm{ExecutionFlowsTimes}_{\pi}(G)\right\}\mid\forall\pi\right\}$.

If the solution is not unique, the choice can be made according to the minimum memory requirement of the robot, which we explain in the next section.

The preceding method only solves the question of reducing the maximum time. If we are interested in reducing the total time, then we must add up all the times of the execution flows with respect to the mapping and find the minimum of this value, i.e., find:
\begin{equation}\label{eq:eq4}
MT=\min\left\{\sum_{P_i\in\mathrm{ExecutionFlows}(G)}T^{\pi}(P_i)\mid\forall\pi\right\},
\end{equation}
and then find the particular mapping $\pi$ that produces the minimal value, i.e., find the set
$$\left\{\pi\mid \sum_{P_i\in\mathrm{ExecutionFlows}(G)}T^{\pi}(P_i)= MT \right\},$$
where $MT$ is obtained from equation \eqref{eq:eq4}.

In the previous formulation, one possibility is that we know in advance at which locations some of the algorithms need to be executed. This prior information has a direct impact on the result. However, in this case, we can reduce the total number of mappings by considering only all mappings that follow the assumption that some of the algorithms must be executed at certain locations.
\begin{remark}[Algebra of Time]\label{Atime}
In the preceding, we have used the algebra of time. For consistency, we give a mathematical formulation.

Let $A_1$ and $A_2$ be two algorithms running on the same platform. Then:
\begin{itemize}
\item if $A_1$ and $A_2$ are in parallel, then $T_X(\{A_1,A_2\})=\max\{T_X(A_1), T_X(A_2)\}$,
\item if $A_1$ and $A_2$ are in serial, then $T_X(\{A_1,A_2\})=T_X(A_1)+ T_X(A_2)$,
\end{itemize}

where $T_X(A_j)$ is the response time of algorithm $A_j$ with $j=1,2$ requested by the edge, fog, and cloud, respectively, for $X=E$, $F$, and $C$.

For more than $2$ algorithms, the algorithms can be viewed as disjoint execution flows (as sub-execution flows of $\mathrm{ExecutionFlows}(G)$ with possible repetition of some of the algorithms) using the associativity of operations defined on both the serial and parallel algorithms. This allows us to describe the algebra of time recursively.
\end{remark}
\subsection{Memory Optimization}
\begin{figure}[t]\centering
\includegraphics[width=0.6\linewidth]{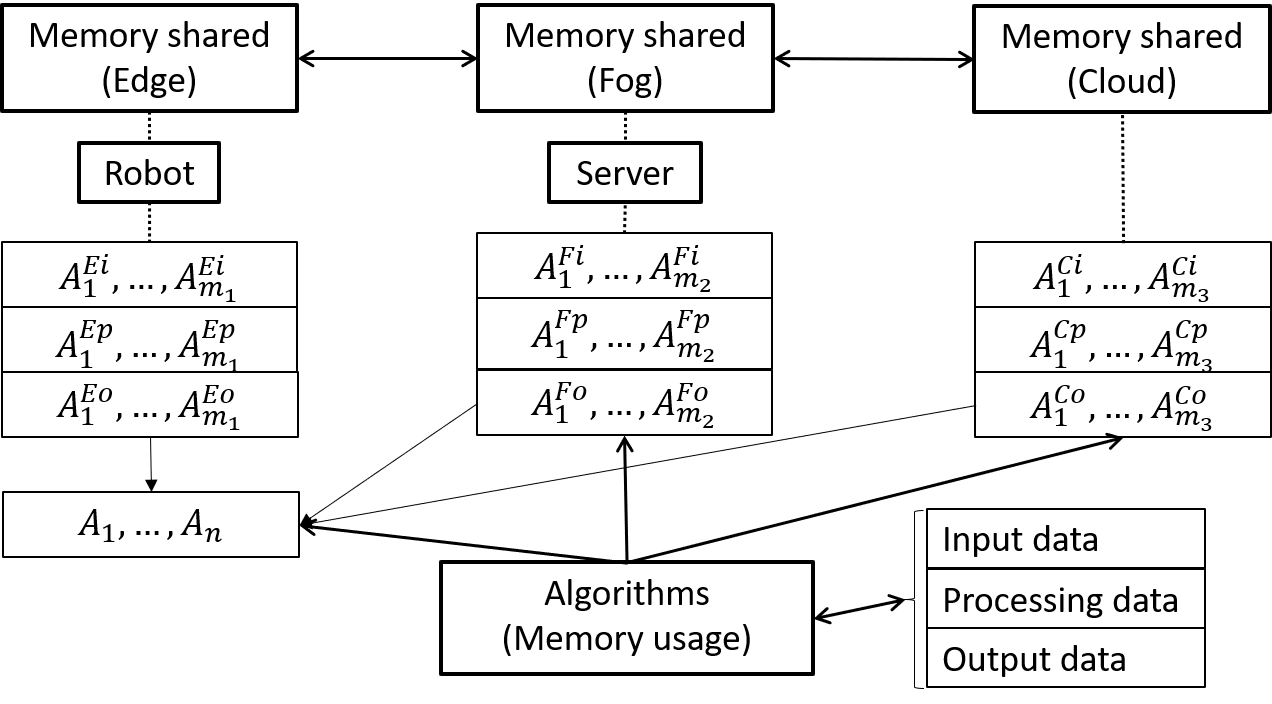}
\caption{Sketch of memory usage. $A^{Ei}_j$, $A^{Ep}_j$, and $A^{Eo}_j$ represent respectively the input data, the processing data, and the output data of the $j$-th algorithm allocated on the edge. The superscripts $E$, $F$, and $C$ represent the location where an algorithm is allocated on the edge, fog, and cloud, respectively.}
\label{fig5}
\end{figure}

There are several studies on shared and distributed memory for algorithms, see \cite{Cheriton:1985, Li:1989, zhou:1990, stumm:1990}. The algorithms share memory and each has access to the specific part of memory where the particular data is stored. The algorithms are allowed to read, write and modify the memory. The necessary memory usage by the outputs of all algorithms, regardless of their allocation, must be taken into account in the robot, see Figure~\ref{fig5}. This is because the output of each algorithm in the set of algorithms for the single robot cloud robotic system, is used by the robot to perform a task.

The data usage of the algorithms has different types:
\begin{itemize}
\item Input data: includes all information needed before starting the process, such as pre-evaluated results, inputs, prior information about the problem to be solved, and so on;
\item Processing data: includes all information needed during processing, such as all variables, data transformations, files to be saved or modified, and so on;
\item Output data: contains all the results of the algorithm.
\end{itemize}
We assume that the outputs of all algorithms must be transmitted to the robot and that the robot then acts accordingly. If we assume that there is an algorithm that is only used as a pre-evaluation of some other algorithms, then the data for that algorithm is stored as the input data of the other algorithms. In Figure~\ref{fig5}, the following set equalities hold:
\begin{align*}
\left\{A_1,\ldots,A_n\right\}&=\left\{A^{Ej}_{1},\ldots,A^{Ej}_{m_1}\right\}\cup\left\{A^{Fj}_{1},\ldots,A^{Fj}_{m_2}\right\}\\
&~~\cup\left\{A^{Cj}_{1},\ldots,A^{Cj}_{m_3}\right\},~~ \text{ for }~~j=i,p,o.
\end{align*}
It is easy to observe that memory usage is a dynamic system where the amount of memory usage changes over time.

\subsubsection{Upper-bound memory estimation for running each algorithm}

Recall that $\mathrm{ExecutionFlows}(G)$ is the set of all execution flows from $\mathbf{1}$ to $\mathbf{0}$ in $\mathcal{ SL }(G)$, where $\mathcal{ SL }(G)$ is shown in Figure~\ref{fig4}.

We classify algorithms according to the order of their occurrence in the execution flows along with their processing memory. Classification is an iterative process that can be done in several steps. In the first step, we start from some initial input data, first compare the space complexities of the algorithms of the first nodes in all execution flows in $\mathrm{ExecutionFlows}(G)$, and define the set $S_1$ as the set of those algorithms with minimal space complexities. Then we drop the algorithms in $S_1$ from all execution flows and consider acceptable execution flows. An acceptable execution flow is an execution flow where all the necessary information for the execution of all algorithms is obtained by executing all the algorithms in $S_1$. The second step is to repeat the first step and find the set $S_2$ as the set of first nodes algorithms in the new set of execution flows with minimal space complexities, and then remove $S_1\cup S_2$ from all execution flows and consider acceptable execution flows. Then iterate in a similar way until all algorithms are removed. Note that in the preceding procedure, by initial algorithms, we mean the set of the first non-virtual algorithm in each execution flow.

Since the number of algorithms is finite, there is a step $z$ such that all execution flows contain no algorithm, in which case the step $z+1$ is just a repetition of the first step, but under the new input data. New input data is basically an update of the initial input data, with the changes obtained by applying all the algorithms in step $1$ to $z$. In this way of defining steps, the steps are defined recursively, and theoretically, an infinite number of steps can be defined.

Note that this way of defining steps allows us to find possible increases in memory usage by algorithms, in the long run, that cannot be seen and evaluated if we assume only finitely many steps.

After identifying steps, in order to observe the memory usage of algorithms, we need to identify the memory usage of algorithms in the long run.

Fix the step $s$ and let $AM^s\_$ be a functor\footnote{A functor is a mapping between categories closed under composition and identity, see \cite{Maclane:1998}.} from the set of algorithms to the total memory used by an algorithm (size of input data plus required execution memory plus size of output data).

To visualize the memory usage of an algorithm, one can plot the values of $AM^s\_$ for that algorithm as a function of steps. If for an algorithm $A_i$, $AM^s A_i$ increases as a function of step $s$, then this implies that we need infinite memory in the infinite run. In this case, we restrict the $AM^s A_i$ to the input data, the processing data, and the output data. Assuming that at least one of the restrictions is unbounded. Therefore, the algorithm $A_i$ must be executed on the cloud due to storage limitations.

Now we only need to consider the algorithms with bounded $AM^s\_$. Note that in the preceding arguments, by unbounded memory usage, we mean that the size of the data used by the algorithm will increase over time.

Following the preceding observations, we can assume, without loss of generality, that all algorithms have bounded $AM^s\_$. In this situation, we can allocate the maximum memory size for the necessary variables and parameters of an algorithm, then the memory size will be fixed, and the only changes will be the values of the variables and parameters.

\subsubsection{Upper-bound memory estimation for running all the algorithms}
Consider the mapping \eqref{eq:eq1}. For $j\in\{E,F,C\}$, define $\pi^{-1}(j)=\left\{A_i\mid\pi(A_i)=j,~~i=1,2,\ldots,n\right\}$.

For each mapping $\pi$, find 
\begin{equation}\label{eq:eq5}
L_{i,j}=\pi^{-1}(j)\cap S_i, 
\end{equation}
with $j=E,F,C$ and $i=1,\ldots,z$. Depending on the mapping, some of the sets $L_{i,j}$ may be empty sets. Note that the set of algorithms within an $L_{i,j}$ cannot be serial by construction.

Regardless of where an algorithm $A_i$ is executed, we consider three types of memory usage, namely $m_{in}(A_i)$, $m_{pr}(A_i)$, and $m_{ou}(A_i)$, which represent the memory requirements for the input data, the processing data, and the output data, respectively. For each step $s$, we have $AM^s A_i\subseteq m_{in}(A_i)\cup m_{pr}(A_i)\cup m_{ou}(A_i)$. After an algorithm is executed, the input and output data are updated if necessary.

For a mapping $\pi$, let $M^E_{\pi}$, $M^F_{\pi}$, and $M^C_{\pi}$ be the maximum amount of memory required on the edge, fog, and cloud, respectively, to correctly execute the corresponding algorithms on them according to the mapping $\pi$.

Now we formulate the algebra of memory usage of algorithms. For clarity, we will later describe an intuition behind the algebraic model for memory. Let $A_1$ and $A_2$ be two algorithms executing on the same platform. Then:
\begin{itemize}
\item if $A_1$ and $A_2$ are in parallel, then:
\begin{align*}
m_{in}(\{A_1,A_2\})=&m_{in}(A_1)\cup m_{in}(A_2),\\
m_{pr}(\{A_1,A_2\})=&m_{pr}(A_1)\cupdot m_{pr}(A_2),\\
m_{ou}(\{A_1,A_2\})=&m_{ou}(A_1)\cup m_{ou}(A_2),
\end{align*}
where $\cupdot$ in the second equation is the disjoint union of sets;
\item if $A_1$ and $A_2$ are in serial, then $m_{in}(\{A_1,A_2\})$ and $m_{ou}(\{A_1,A_2\})$ can be obtained as in the parallel case, but $m_{pr}(\{A_1,A_2\})=m_{pr}(A_j)$, where $m_{pr}(A_j)$ is chosen in the second equation with $j=1,2$ such that $|m_{pr}(A_j)|=\max\left\{|m_{pr}(A_1)|, |m_{pr}(A_2)|\right\}$,
where $|m_{p}(A_j)|$, denotes the size of the memory.
\end{itemize}
For more than $2$ algorithms, the algorithms can be viewed as disjoint execution flows (as sub-execution flows of $\mathrm{ExecutionFlows}(G)$ with possible repetition of some of the algorithms) using the associativity of operations defined on the parallel algorithms and also on the serial algorithms. This allows us to recursively describe the algebra of memory usage of algorithms.

For a mapping $\pi$ to obtain $M^E_{\pi}$, $M^F_{\pi}$, and $M^C_{\pi}$, we first apply the operations $m_{in}$, $m_{pr}$, and $m_{ou}$ to entries of the matrix $L=\left(L_{i,j}\right)_{z\times3}$ in Equation \eqref{eq:eq5}. That is $M_{X}(L)=\left(m_{X}(L_{i,j})\right)_{z\times3}$, with $X\in\{in,pr,ou\}$. Then we have:
\begin{align*}
M^k_{\pi}=&\left|\bigcup_{i=1}^z\left(m_{in}(L_{i,k})\cup m_{ou}(L_{i,k})\right)\right|+\left|\bigcupdot_{i=1}^z\left(m_{pr}(L_{i,k})\right)\right|,
\end{align*}
where $k\in\{E,F,C\}$ and the absolute is to denote the size of the memory. The most important one is $M^E_{\pi}$, the memory usage of the robot. The solution is obtained by determining a bound that specifies the set of all possible mappings (solutions) such that the size of $M^E_{\pi}$ is smaller than this bound. Note that from the last component of $M^E_{\pi}$, we can find a reasonably small upper bound for the required processing memory of the robot. The least upper bound for the required processing memory of the robot can only be found by a deep understanding of all algorithms and their components.

\begin{remark}[Algebra of Memory Usage]\label{Amemory}
Note that the memory usage of each algorithm is a triple $(m_{in},m_{pr},m_{ou})$, where $m_{in}$ is the total memory used as input, $m_{pr}$ is the total memory used during processing, and $m_{ou}$ is the total memory used as output of the algorithm $a_i$. Moreover, the set of $m_{pr}$'s for all algorithms is assumed to be nested, i.e., if the required processing memory of algorithm $a$ is smaller than the required processing memory of algorithm $b$, then $m_{pr}(a)\subset m_{pr}(b)$. The algebra of memory usage can be visualized in Figures~\ref{fig5p} and \ref{fig6p}.

\begin{figure}[t]\centering
\includegraphics[width=0.5\linewidth]{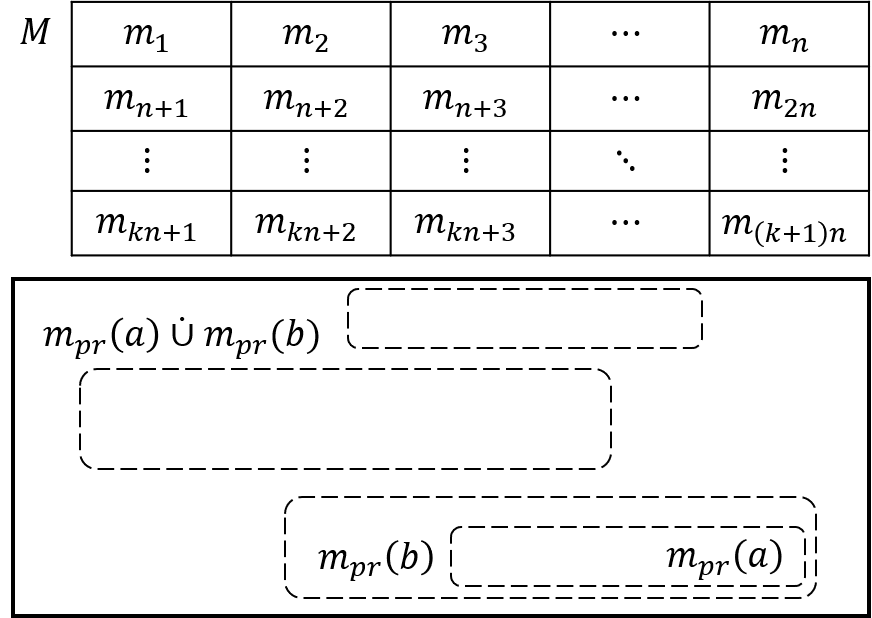}
\caption{Algebra of memory usage. $\dot{\cup}$ represents the disjoint union.}
\label{fig5p}
\end{figure}

In Figure~\ref{fig5p} the following statements hold:
\begin{itemize}
\item $m(a)=\Big(m_{in}(a)=\{m_{\alpha_1},\ldots,m_{\alpha_{\delta_1}}\}, m_{pr}(a)=\{m_{\alpha'_1},\ldots,m_{\alpha'_{\delta_2}}\}, m_{ou}(a)=\{m_{\alpha''_1},\ldots,m_{\alpha''_{\delta_3}}\}\Big)$;
\item $m(b)=\Big(m_{in}(b)=\{m_{\beta_1},\ldots,m_{\beta_{\gamma_1}}\}, m_{pr}(b)=\{m_{\beta'_1},\ldots,m_{\beta'_{\gamma_2}}\}, m_{ou}(b)=\{m_{\beta'_1},\ldots,m_{\beta'_{\gamma_3}}\}\Big)$;
\item either $m_{pr}(a)\subseteq m_{pr}(b)$ or $m_{pr}(b)\subseteq m_{pr}(a)$ and we assume $m_{pr}(a)\subseteq m_{pr}(b)$;
\item if algorithms $a$ and $b$ are parallel, then:
$$\begin{array}{l}
m(a \wedge b)=\Big (m_{in}(a)\cup m_{in}(b),m_{pr}(a)\cupdot m_{pr}(b)\\
~~~~~~~~~~~~~~~,m_{ou}(a)\cup m_{ou}(b)\Big),
\end{array}$$
where $\cupdot$ stands for disjoint union and $m(a \wedge b)$ is the total memory usage for the executions of the two algorithms $a$ and $b$;
\item if algorithms $a$ and $b$ are serial, then $m(a \wedge b)=\Big(m_{in}(a)\cup m_{in}(b),m_{pr}(b),m_{ou}(a)\cup m_{ou}(b)\Big)$.
\end{itemize}

\begin{figure}[t]\centering
\includegraphics[width=0.5\linewidth]{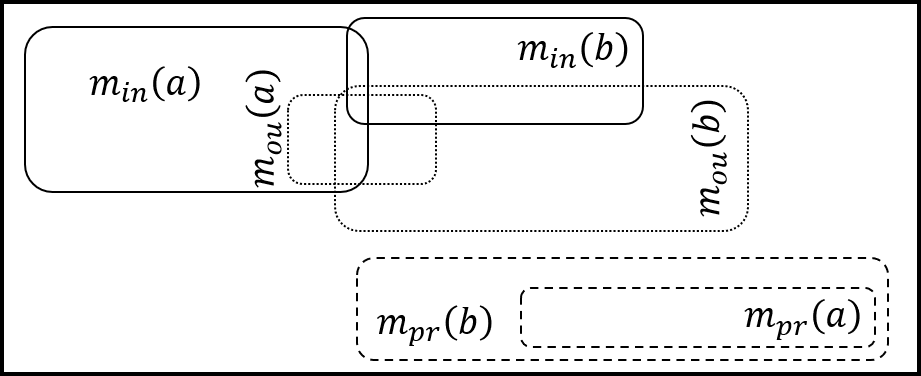}
\caption{General intuition for $m_{in}$, $m_{ou}$, and $m_{pr}$ of two algorithms $a$ and $b$. In this figure, $m(a)=(m_{in}(a),m_{pr}(a),m_{ou}(a))$ and $m(b)=(m_{in}(b),m_{pr}(b),m_{ou}(b))$ are the triples for the memory uses by algorithms $a$ and $b$ under the assumption that $m_{pr}(a)\subseteq m_{pr}(b)$.}
\label{fig6p}
\end{figure}
\end{remark}

\subsection{Memory-Time Optimization}
\begin{figure}[t]\centering
\includegraphics[width=0.5\linewidth]{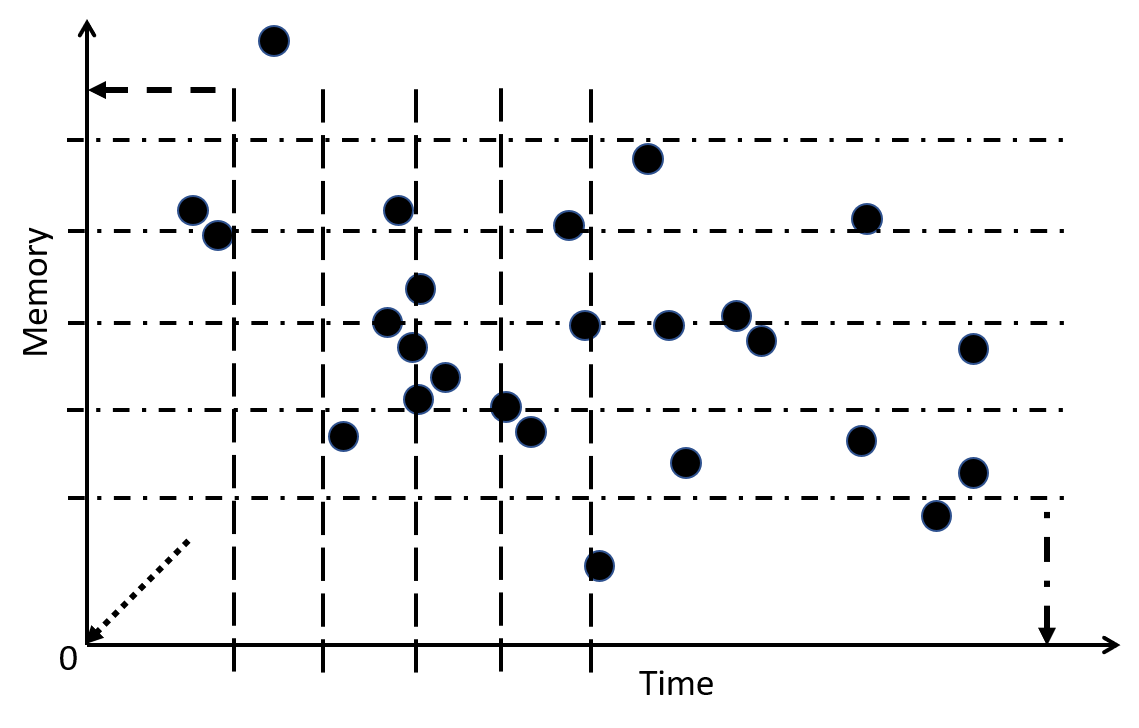}
\caption{The two-dimensional space consists of time and memory components. Each point is the memory and time with respect to a particular solution. An optimized solution for time and memory components for distributing algorithms can be obtained by minimizing the distance to the axis's or center point, $0$. Each black circle represents the memory and time for a particular solution, the dotted arrow shows the direction to the optimal solution with respect to time and memory, the dashed arrow shows the direction to the optimal solution with respect to time, and the dash-dotted arrow shows the direction to the optimal solution with respect to memory.}
\label{fig6}
\end{figure}

The problem of optimizing the memory capacity and the overall time is solved by finding the list of solutions for the time component together with the list of solutions for the memory component, and then depending on the component we are interested in, we find the solution using the method described below. Figure~\ref{fig6} contains an illustration of the results we follow. Note that for the list of solutions, we only consider the solutions that are obtained from the same mapping $\pi$ in \eqref{eq:eq1}.

\section{Method}

Our method can be explained as follows:
\begin{itemize}
\item[] \textbf{Input:}
\begin{enumerate}
\item List of algorithms $\{A_1,\ldots,A_n\}$; 
\item Execution dependency of algorithms; 
\item Execution time of algorithms on the cloud, fog, and edge; 
\item Data transmission speed (time) between cloud, fog, and edge;
\item Input, processing and output memory size of algorithms;
\item Additional information about where to run the algorithms. (optional)
\end{enumerate}
\item[] \textbf{Output:} The mapping $\pi:\{A_1,\ldots,A_n\}\rightarrow\{E,F,C\}$ for algorithm allocation such that it gives the optimal overall time and memory allocation.
\item[] \textbf{Steps:} The following steps should be followed:
\begin{enumerate}
\item Construct the graph of algorithms, $G$, and its corresponding semi-lattice $\mathcal{ SL }(G)$; 
\item Find the set of all execution flows, $\mathrm{ExecutionFlows}(G)$;
\item Make a guess about the optimal allocation algorithms and find its overall time and memory;
\item Apply the branch and bound algorithm to the elements of $\mathrm{ExecutionFlows}(G)$, one at a time as follows. Note that allocations of $\mathbf{1}$ and $\mathbf{0}$ are on the edge, and for the other algorithms, they are on a subset of $\{E,F,C\}$ (the subset because with the prior information about the algorithms, we may be able to remove at least one of $E$, $F$, and $C$):
\begin{enumerate}
\item Apply the algebra of memory~\ref{Amemory} and the algebra of time~\ref{Atime} to the subterms of $\mathrm{ExecutionFlows}(G)$ induced by considering the algorithms already allocated in the previous steps and find the partial overall memory and time.
\item Compare the results of the preceding step with the guessed optimal solution. If the distance to the origin is greater, then stop the branching, and if it is less or equal, proceed to the next algorithm;
\item If all algorithms are allocated, and the distance to the origin is less than the guessed optimal solution, then update the guess about the optimal allocation algorithms with the current overall time and memory and proceed to the next possible branch.
\end{enumerate}
\item The updated guess about the optimal allocation after completing the preceding steps is the optimal allocation.
\end{enumerate}
\end{itemize}

\section{Experiments}
To allow a comparison between our result and the result obtained in \cite{li:2018}, we first give two very simple but detailed examples and then compare them with more complex examples. Example~\ref{exmpl1}, shows the importance of the data transmission time on the overall performance by comparing our result and the result obtained by the method proposed in \cite{li:2018}.
\begin{example}\label{exmpl1}
Suppose a robot is given a list of numbers to sort. Note that the authors in \cite{li:2018} did not consider fog, but it is easy to add it to their method. Assume that the robot can sort the list in $5$ seconds, the cloud processor and the fog processor are $5$ and $1.5$ times faster than the robot, respectively, and suppose that the list can be transferred to the cloud in $1.5$ and to the fog in $3$ seconds (with no delay), then in the formulation in \cite{li:2018} we have:
$$t^e=\left\{\begin{array}{ll}
t^s+R_c&=3+1,\\
t^s+R_f&\approx1.5+3.33,\\
t^s+R_r&=0+5,
\end{array}\right.$$
where, as indicated in \cite{li:2018}, $t^e$ denotes the algorithm end time, $t^s$ denotes the algorithm start time, and $R_x$ denotes the execution time of an algorithm executed on the cloud, fog, and edge for $x=c$, $x=f$, and $x=r$, respectively. Therefore, the $\min$ can be obtained from the execution of the sorting algorithm on the cloud. But then it is necessary to return the result of sorting to the robot, which requires additional $3$ seconds. This means that the overall performance of this system is better when the sorting algorithm is executed on the cloud (by \cite{li:2018}) and on the edge (by ours). Therefore, the overall time is $7$ seconds when we consider the result of algorithm allocation by \cite{li:2018} and $5$ seconds when we consider our algorithm allocation. Even if we reduce the data transmissions by a factor of $3$ in this example, nothing changes in the result of \cite{li:2018}, but our result changes because the overall performance of this system is now better when the algorithm is run in the cloud. For the dependence on data transmission speed, see Example~\ref{exmpl2}.
\end{example}

In Example~\ref{exmpl2}, we present a detailed comparison between our result and the result obtained in \cite{li:2018} for all possible values for the data transmission time, and we highlight the conditions for the task allocations.
\begin{example}\label{exmpl2}
Here we consider an example similar to Example~\ref{exmpl1}. Suppose that a robot is given a list of numbers to sort. Assume that the robot can sort the list in $5$, the cloud in $1$, and the fog in $2$ seconds, and assume that the list can be transmitted to the cloud in $2x$ seconds (without delay) and to the fog in $x$ seconds. Then, depending on the value of $x$, the optimal allocation of the algorithm is obtained and shown in Table~\ref{tab1}.

\begin{table}[t]
\caption{Optimal allocation of the sort algorithm. $x$ is the data transmission time from fog to edge and from fog to cloud.}
\begin{center}
\begin{tabular}{l|cc}
&Solution by \cite{li:2018}& Ours\\
\hline
Cloud&$x<1$&$x<\frac{1}{2}$\\
Fog&$1<x<3$&$\frac{1}{2}<x<\frac{3}{2}$\\
Edge&$x>3$&$x>\frac{3}{2}$
\end{tabular}
\end{center}
\label{tab1}
\end{table}%
We show in Figure~\ref{fig1p} the overall time from the moment the robot sends the request to execute the algorithm to the moment the robot has the result of the algorithm. Figure~\ref{fig1p} shows that our method better describes the overall time for algorithm allocation. As can be seen in this figure, there are certain intervals where the overall time for our optimal algorithm allocation differs from the optimal algorithm allocation proposed by \cite{li:2018}.

\begin{figure}[t]\centering
\includegraphics[width=0.6\linewidth]{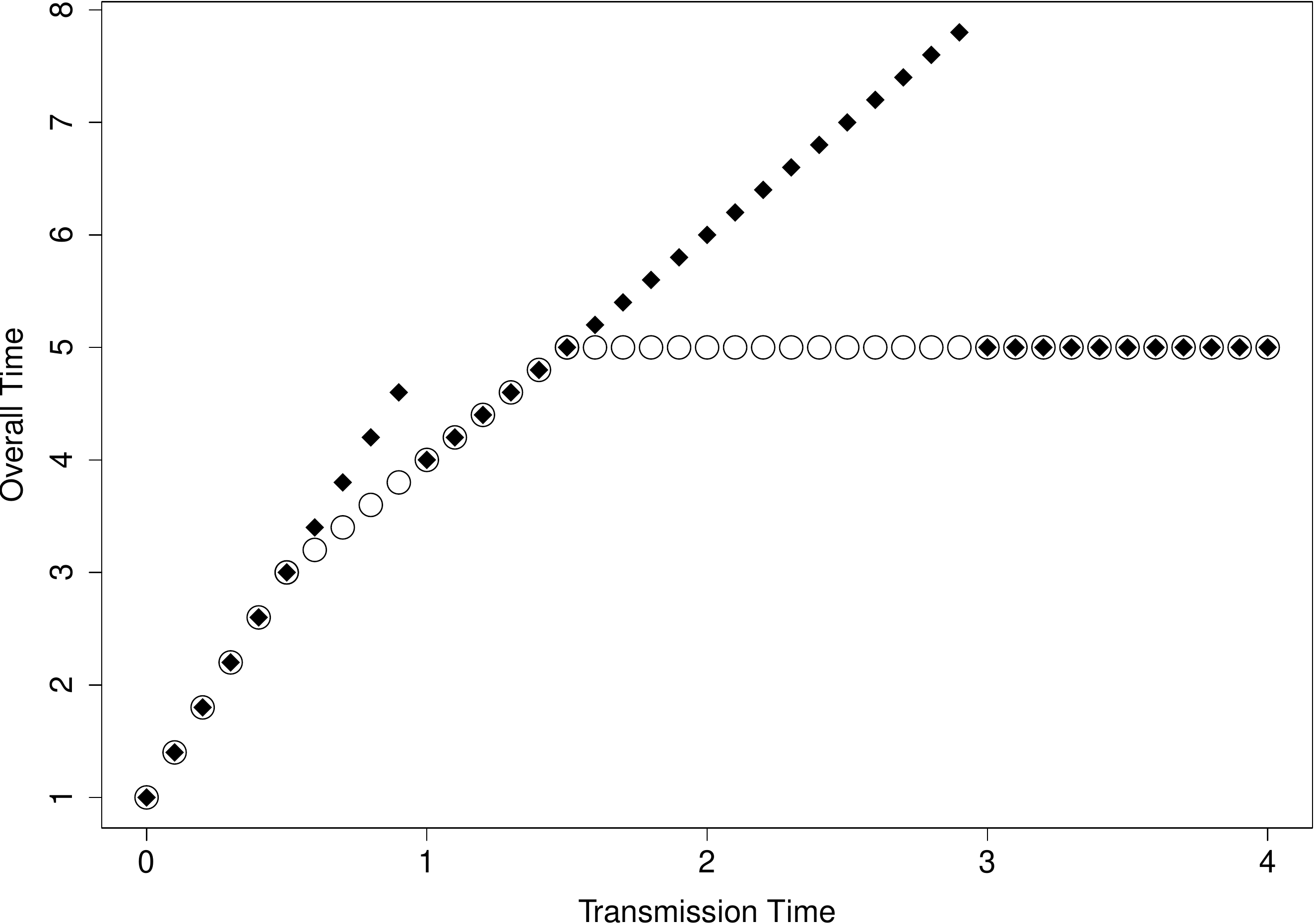}
\caption{The overall time from the moment the robot sends the request to execute the algorithm to the moment the robot has the result of the algorithm. We compare the time obtained with our method, circles, and the time obtained with \cite{li:2018}, diamonds.}
\label{fig1p}
\end{figure}
\end{example}
Examples~\ref{exmpl1} and \ref{exmpl2} show that when using the result of our method, the overall time is always less than or equal to the overall time of the result of the method proposed by \cite{li:2018}. A clear difference in the overall time is shown in Figure~\ref{fig1p}.

We have tested the proposed model and compare it with the method of\cite{li:2018}. 

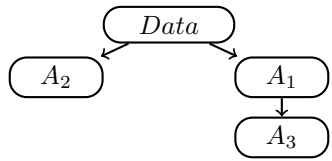
\begin{figure}[t]\centering
\begin{tikzpicture}
\begin{scope}
    \node (A) at (0,0) [text width=1.5cm,rectangle,thick,draw,rounded corners=1.5ex,align=center] {$Data$};
    \node (B) at (1.5,-0.7)  [text width=1cm,rectangle,thick,draw,rounded corners=1.5ex,align=center] {$A_1$};
    \node (C) at (-1.5,-0.7)  [text width=1cm,rectangle,thick,draw,rounded corners=1.5ex,align=center] {$A_2$};
    \node (D) at (1.5,-1.5) [text width=1cm,rectangle,thick,draw,rounded corners=1.5ex,align=center]{$A_3$};
\end{scope}
    \path [->] (A) edge[thick,->] (B);
    \path [->] (B) edge[thick,->] (D);
    \path [->](A) edge [thick,->] (C);
\end{tikzpicture}
\caption{Algorithms $A_1$ and $A_2$ read the dataset and are executed in parallel. Algorithm $A_3$ cannot be executed until algorithm $A_1$ has finished.}
\label{fig7}
\end{figure}

We performed a simulation for a cloud robotic system consisting of a single robot, a fog, a cloud, and three algorithms to be performed with a fixed dataset (input data). The corresponding graph of the algorithms can be seen in Figure~\ref{fig7}.

For the memory usage of the dataset, we assume a maximum size of $500$ $MegaBytes$, and the algorithms $A_1$, $A_2$, and $A_3$ require $300$, $50$, and $100$ $MegaBytes$ respectively, and also with different space complexities $O(n^2)$, $O(n)$, and $O(n \log(n))$ respectively. The average execution time of the algorithms depends on where they are executed and is given in Table~\ref{tab0}.
\begin{table}[t]
\caption{Average execution time of all algorithms depending on their execution location, in seconds.}
\begin{center}
\begin{tabular}{c|cccc}
&Dataset&$A_1$&$A_2$&$A_3$\\
\hline
Edge&0&2&4&6\\
Fog&0&1&2&3\\
Cloud&0&0.5&1&1.5
\end{tabular}
\end{center}
\label{tab0}
\end{table}
We consider all possible mappings 
$\pi:\{dataset,A_1,A_2,A_3\}\rightarrow\{E,F,C\}$.

\begin{figure}[t]\centering
\includegraphics[width=0.6\linewidth]{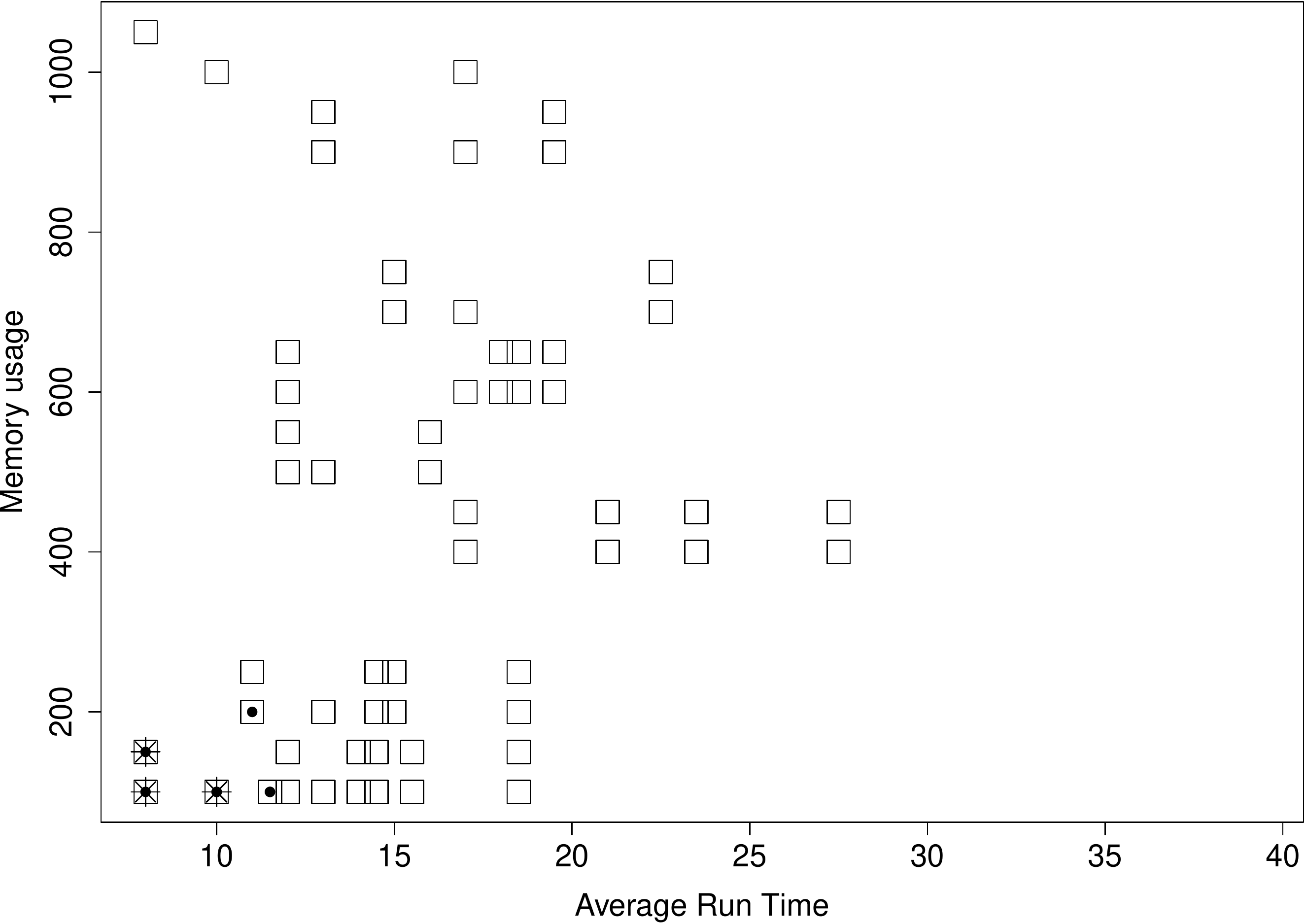}
\caption{Memory-Time relationship assuming that the communication time is fixed. Also, the minimal Memory-Time values in the Memory-Time relation are identified with the circles under the assumption of fixed communication time. The asterisks correspond to the points with minimum values of Memory-Time in the Memory-Time relation, assuming an additional communication delay. The squares are all possible Memory-Time under the assumption without additional communication delay.}
\label{fig10}
\end{figure}

\begin{figure}[t]\centering
\includegraphics[width=0.6\linewidth]{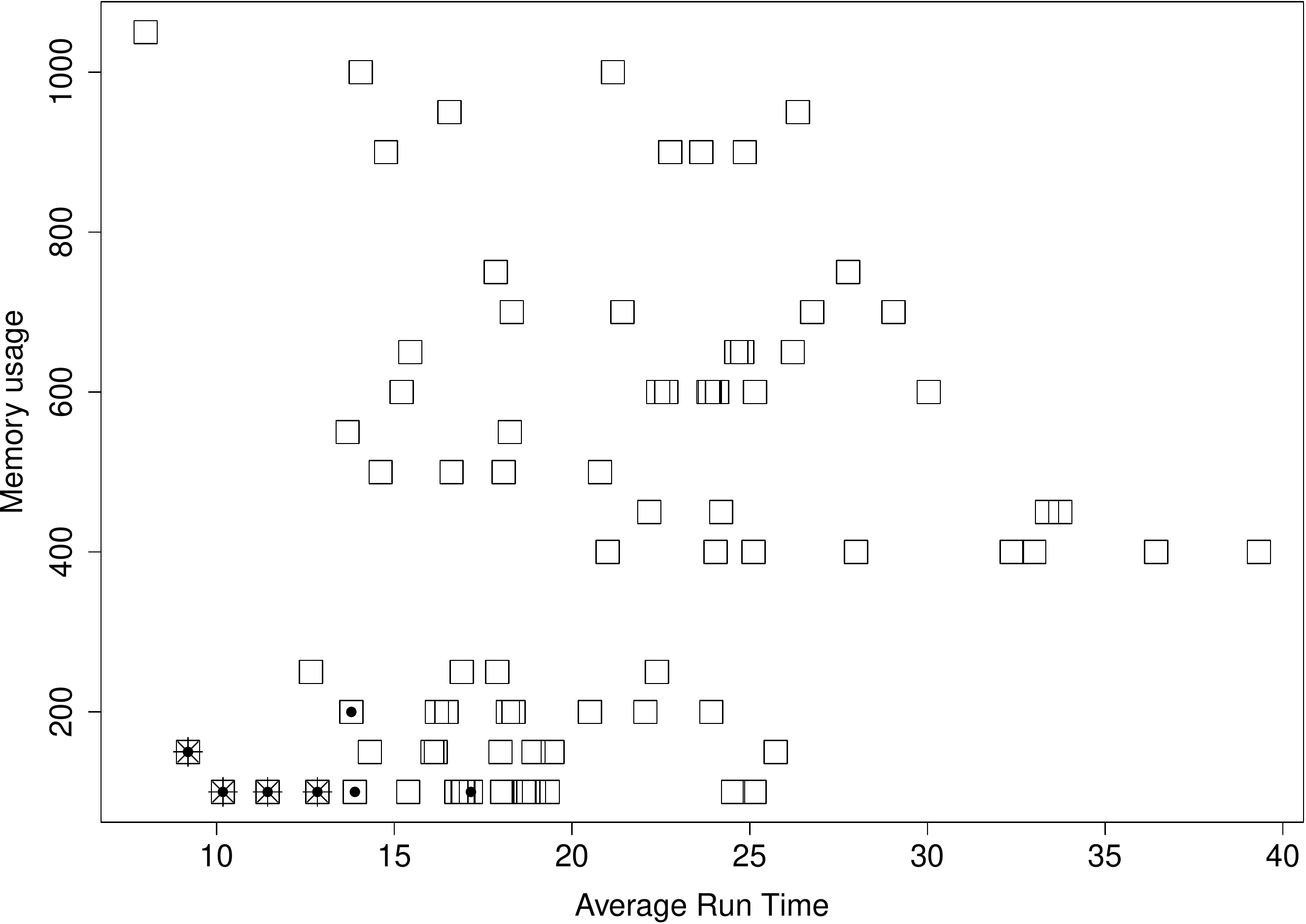}
\caption{ Memory-Time relationship assuming an additional communication delay. Similar to Figure~\ref{fig10}, but swapping the assumption from ''with fixed communication time'' to ''with additional communication delay''.}
\label{fig11}
\end{figure}

For simplicity, we first assume that the communication is done with a fixed time, $2$ seconds data transfer time between fog and edge, and fog and cloud. Figure~\ref{fig10} shows the relation Memory-Time with respect to the mapping $\pi$ for the fixed communication time. We also study a more realistic scenario, namely the case with an additional communication delay. It consists of the absolute value of a random variable with normal distribution $\mathcal{N}(0,1)$. Figure~\ref{fig11} shows the Memory-Time relation with respect to the mapping $\pi$ with additional communication delay.

The minimum values of Memory-Time on a fixed interval of size $d=350$ are shown in Figure~\ref{fig10}. To better describe the optimality in time and memory of our algorithms allocation, we plotted the values of memory and time with respect to all possible allocations of algorithms and then highlighted the solutions.

In Figure~\ref{fig11}, similar to Figure~\ref{fig10}, we identify the same points with minimal Memory-Time values on a fixed interval of size $d=350$, but in Memory-Time relation assuming an existing additional communication delay. 

Note that the points in Figures~\ref{fig10} and \ref{fig11} show the average execution time of all algorithms by the total memory usage of the robot with respect to a mapping $\pi$. This means that each point also contains information about where each algorithm needs to be executed. This is essentially done by defining an order (LEX-order) on the set of mappings and assigning each point to that order.

Table~\ref{tab2} shows the corresponding placement of the algorithms and the dataset assuming the fixed communication time, and these places provide the minimum values of Memory-Time on a fixed interval of size $d=350$. The minimum value is shown in the first row.

\begin{table}[t]
\caption{Allocation of algorithms and dataset that have minimum total cost assuming fixed communication time.}
\begin{center}
\begin{tabular}{ccccc}
Dataset&$A_1$&$A_2$&$A_3$&Distance to Origin\\
\hline
Fog&Fog&Fog&Fog&\textbf{1131.4}\\
Fog&Fog&Edge&Fog&1159.3\\
Cloud&Cloud&Cloud&Cloud&1305.4\\
Cloud&Cloud&Fog&Cloud&1305.4\\
Cloud&Cloud&Cloud&Fog&1436.9\\
Cloud&Cloud&Fog&Fog&1436.9\\
Fog&Fog&Fog&Edge&1442.2
\end{tabular}
\end{center}
\label{tab2}
\end{table}

In Table~\ref{tab3}, we show the corresponding placement of the algorithms and the dataset assuming the additional communication delay, and these places provide the minimum values of Memory-Time on a fixed interval of size $d=350$. The minimum value is shown in the first row.

\begin{table}[t]
\caption{Allocation of algorithms and dataset that have minimum total cost assuming the additional communication delay.}
\begin{center}
\begin{tabular}{ccccc}
Dataset&$A_1$&$A_2$&$A_3$&Distance to Origin\\
\hline
Fog&Fog&Fog&Fog&\textbf{1205.5}\\
Fog&Fog&Edge&Fog&1236.9\\
Cloud&Cloud&Cloud&Cloud&1384.6\\
Cloud&Cloud&Fog&Cloud&1550.5
\end{tabular}
\end{center}
\label{tab3}
\end{table}
\begin{remark}
In Figures~\ref{fig10} and \ref{fig11}, we have shown the memory usage of the robot and the average running time of all algorithms for all possible ways of allocation of algorithms, which means that the corresponding memory usage of the robot and the average running time of all algorithms for each random algorithm allocation is one of the squares. Therefore, our method provides the optimal algorithm allocation that minimizes the memory usage of the robot and the average running time of all algorithms.
\end{remark}

To make a more realistic comparison between our result and the result obtained in \cite{li:2018}, we assume that in the preceding simulation, the data transmission time between fog and edge, and fog and cloud varies by $1$, $2$, $4$, and $6$ seconds. For the given data transmission times equal to $1$, $2$, $4$, and $6$, the minimum total costs for both methods are shown in Table~\ref{tab4}. To compare the two results, we assume that the data transmission is fixed. As can be seen in Table~\ref{tab4}, the difference in algorithm allocation for data transmission time is equal to $2$ and $4$, while the other method gives the same results. In Table~\ref{tab4}, we only consider the time parameter and ignore the memory parameter. The minimum average time column, where the relative data transmission time is $2$ and $4$ seconds, shows that the overall algorithm allocation time of \cite{li:2018} is higher than ours.
\begin{table}[t]
\caption{Comparison of the allocation of algorithms and dataset having minimum total cost by our method and by \cite{li:2018}, among the four data transmission times equal to $1$, $2$, $4$, and $6$ seconds between fog and edge, and fog and cloud. The $(+x)$ in the table denotes the amount of time $x$ seconds required to send the output of the algorithms to the robot after execution. This does not take into account the memory usage of the robot.}
\begin{center}
\begin{tabular}{l|lcc}
&Allocation&Minimum&Data Transmission\\
&Method&Average Time& Time\\
\hline
Ours&All to Cloud&6&1\\
Ours&All to Fog&8&2\\
Ours&All to Edge&8&4\\
Ours&All to Edge&8&6\\
\cite{li:2018}&All to Cloud&4(+2)&1\\
\cite{li:2018}&All to Cloud&6(+4)&2\\
\cite{li:2018}&All to Fog&8(+4)&4\\
\cite{li:2018}&All to Edge&8&6
\end{tabular}
\end{center}
\label{tab4}
\end{table}

Minimizing overall time is as important as minimizing memory, and it is not enough to minimize either of them and describe the performance of the system using only this information. As can be seen in Figures~\ref{fig10} and \ref{fig11}, if we focus only on minimizing the memory usage of the robot, we ignore the overall execution time, and therefore the system is more time- consuming, and if we only focus on minimizing the time, the robot may not have enough memory and hence may not be able to execute some tasks.

Now we apply our memory usage optimization to both methods. For the cases with the average data transmission equal to $1$ or more than or equal to $4$ seconds, the optimal allocation for both our method and by \cite{li:2018} are to allocate all algorithms to the cloud and to the fog, respectively. But in the cases with the average data transmission equal to $2$ or $4$ seconds, the optimal allocation will change. The optimal allocations with our method are to allocate all algorithms to the fog for both cases, and the optimal allocations using the method by \cite{li:2018} are to allocate all algorithms to the cloud and to the fog respectively for data transmission equal to $2$ and $4$ seconds.

\begin{figure}[t]\centering
\includegraphics[width=0.5\linewidth]{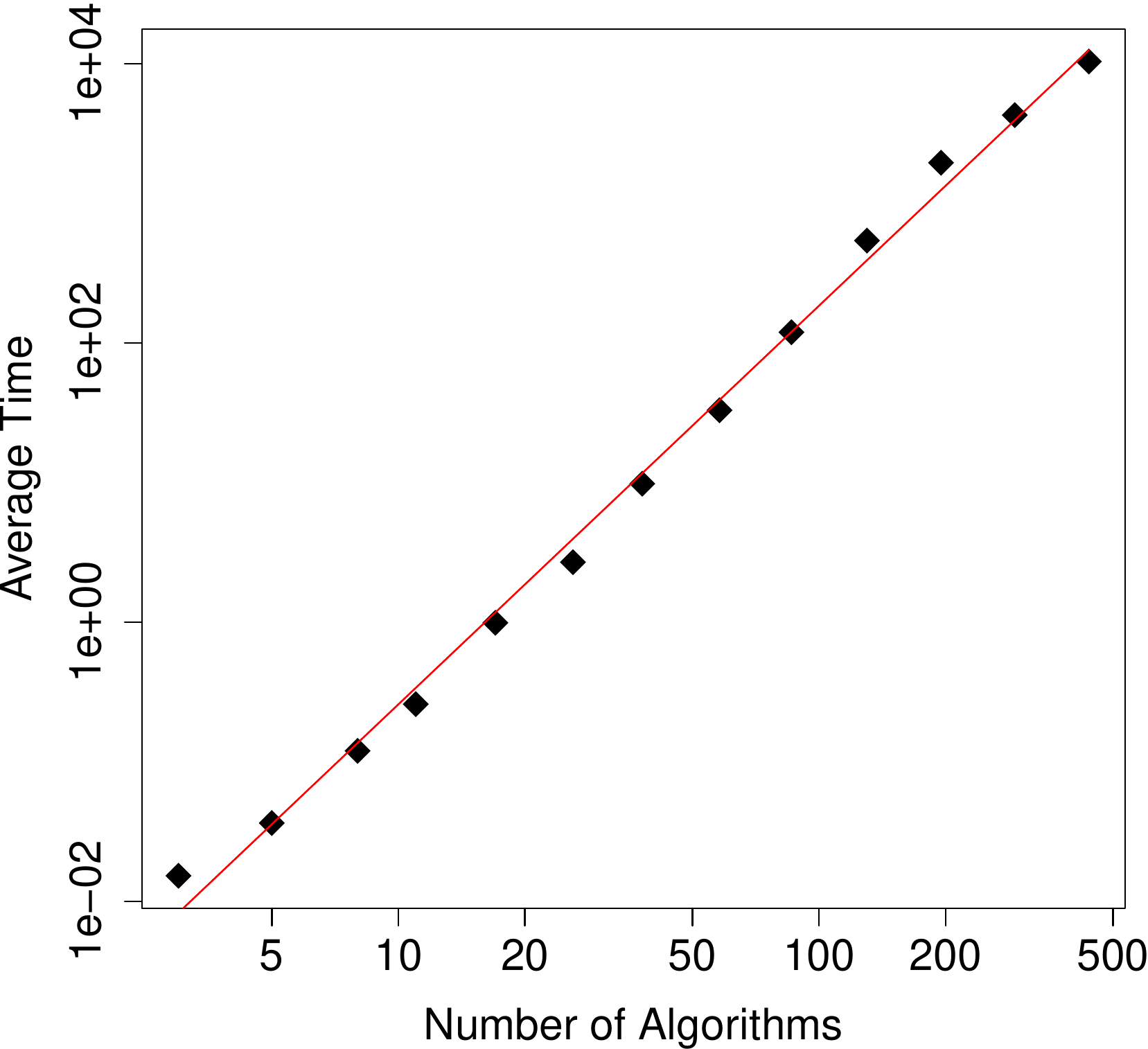}
\caption{Evaluation of scalability. Average time (in seconds) to find an optimal allocation as a function of the number of algorithms, in logarithmic scale (both axis). Black diamonds are the real values and the red line is the fit obtained by the linear regression and its $R^2=0.9958$.}
\label{fig}
\end{figure}

To evaluate the scalability of our method, we conducted an experiment where we randomly generated the graph of all algorithms for a given number of algorithms and plot the average time to find a solution for 10 repetitions for each number of algorithms in Figure \ref{fig}. The figure has both axis in logarithmic scale and shows a linear relation between the average time and the number of algorithms to allocate, hence the time complexity of our method is polynomial.

For a more complex scenario, we test our method on real-world data. We collected real-world data and use it as input parameters in simulations. We first describe the data and then apply our method to the data.

The edge node is a Raspberry Pi 4 Model B. The fog nodes are computers with 32GB of RAM, Intel(R) Core(TM) i7-9700K CPU @ 3.60GHz with 8 cores and an Nvidia GeForce RTX 2080 GPU. The cloud nodes are virtual machines with 64GB of RAM, Intel(R) Xeon(R) Silver 4114 CPU @ 2.20GHz with 2 cores, and an Nvidia Tesla V100 PCIe 16GB GPU. For the architecture, we consider a single edge node with three fog nodes and two cloud nodes. All nodes are considered communicating with each other, and their transmission times are shown in Table~\ref{tab2p}.

Algorithm $A_1$ reads a folder of images of a person's face. The algorithm extracts features from each image and creates a database that associates a person with the features of their face. The second algorithm, $A_2$, stores the database created by its predecessor in a file on disk. $A_3$ loads the database file back into memory so that operations can be performed on it. $A_4$ receives a compressed image of a person's face and decompresses it. $A_5$ extracts facial features from an input image. $A_6$ compares the features of a single image with all images in the database. $A_7$ receives the result of the matching algorithm and identifies the person by finding the image with the most matches.

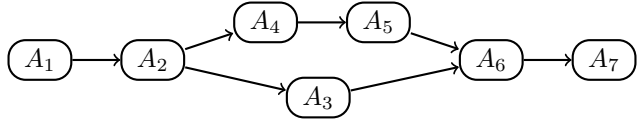
\begin{figure}[t]\centering
\begin{tikzpicture}
\begin{scope}
    \node (A) at (0,0) [text width=0.6cm,rectangle,thick,draw,rounded corners=1.5ex,align=center] {$A_1$};
    \node (B) at (1.5,-0)  [text width=0.6cm,rectangle,thick,draw,rounded corners=1.5ex,align=center] {$A_2$};
    \node (C) at (3.7,-0.5)  [text width=0.6cm,rectangle,thick,draw,rounded corners=1.5ex,align=center] {$A_3$};
    \node (D) at (3,0.5) [text width=0.6cm,rectangle,thick,draw,rounded corners=1.5ex,align=center]{$A_4$};
    \node (E) at (4.5,0.5)[text width=0.6cm,rectangle,thick,draw,rounded corners=1.5ex,align=center] {$A_5$};
    \node (F) at (6,0) [text width=0.6cm,rectangle,thick,draw,rounded corners=1.5ex,align=center]{$A_6$};
    \node (G) at (7.5,0)[text width=0.6cm,rectangle,thick,draw,rounded corners=1.5ex,align=center] {$A_7$};

\end{scope}

    \path [->] (A) edge[thick,->] (B);
    \path [->] (B) edge[thick,->] (C);
    \path [->](B) edge [thick,->] (D);
    \path [->](C) edge [thick,->] (F);
    \path [->](D) edge [thick,->] (E);
    \path [->](E) edge [thick,->] (F);
    \path [->](F) edge [thick,->] (G);
\end{tikzpicture}
\caption{Graph of all algorithms.}
\label{fig1pp}
\end{figure}

The graph of all algorithms is shown in Figure~\ref{fig1pp} (the flow of algorithms is shown from left to right).
For the communication time between nodes for transferring $32$ bytes of data, we have Table~\ref{tab2p}.
\begin{table}[t]
\caption{Transmission time of $32$ bytes of data, in seconds. $\mathcal{ FN }(\mu,\sigma)$ is used to denote the folded normal distribution with parameter $\mu$ and standard deviation $\sigma$. $F$ is used to denote any fog, $C$ is used for any cloud, and $E$ is used to represent the edge node.}\label{tab2p}
\begin{center}
\begin{tabular}{l|l}
Transmission&Time (in seconds)\\
\hline
$C\rightarrow F$&$0.439+\mathcal{FN}(0.188,0.087)$\\
$F\rightarrow C$&$0.417+\mathcal{FN}(0.367,0.365)$\\
$F\rightarrow E$&$0.475+\mathcal{FN}(0.187,0.397)$\\
$E\rightarrow F$&$0.447+\mathcal{FN}(0.182,0.111)$\\
$C\rightarrow E$&$0.116+\mathcal{FN}(0.182,0.024)$\\
$E\rightarrow C$&$0.175+\mathcal{FN}(0.188,0.015)$\\
$C\rightarrow C$&$0.112+\mathcal{FN}(0.030,0.018)$\\
$F\rightarrow F$&$0.115+\mathcal{FN}(0.047,0.025)$
\end{tabular}
\end{center}
\end{table}
Table~\ref{tab3p} shows all algorithms' average execution times to be executed by an edge node, fog, or cloud. Also, the input, output, and processing memory and space complexities of all the algorithms are shown.
\begin{table*}[t]
\caption{Average execution time of algorithms (in seconds) on an edge node, fog, or cloud, along with their input (in bits), output (in bits), processing memory (in bytes), and their space complexities. $n$ is the number of images and $m$ is the size of images.}\label{tab3p}
\begin{center}
\begin{tabular}{l|ccccccc}
&$Edge$ (s)&$Fog$ (s)&$Cloud$ (s)&Input&Output&Processing&Space\\
&&&&Size (bits)&Size (bits)&Size (bytes)&Complexity\\
\hline
$A_1$&$0.445$&$0.153$&$0.047$&$4718592$&$1120$&$14619367$&$O(nm)$\\
$A_2$&$4.475$&$1.538$&$0.470$&$47185920$&$11200$&$11683901$&$O(n)$\\
$A_3$&$7.2\times10^{-4}$&$4.1\times10^{-4}$&$1.5\times10^{-4}$&$11200$&$11200$&$11684220$&$O(n)$\\
$A_4$&$2.0\times10^{-4}$&$7.74\times10^{-5}$&$3.46\times10^{-5}$&$11200$&$0$&$7799083$&$O(m)$\\
$A_5$&$6.61\times10^{-5}$&$1.94\times10^{-5}$&$9.96\times10^{-6}$&$11200$&$11200$&$11253700$&$O(m)$\\
$A_6$&$2.1\times10^{-4}$&$1.3\times10^{-4}$&$4.75\times10^{-5}$&$11200$&$1120$&$11261700$&$O(nm)$\\
$A_7$&$1.09\times10^{-3}$&$4.01\times10^{-3}$&$2.7\times10^{-4}$&$4718592$&$4718592$&$8010779$&$O(n)$
\end{tabular}
\end{center}
\end{table*}
By applying our method, it is found that to achieve the optimal algorithm allocation, all algorithms should be allocated to one of the fog nodes depending on the communication delays, and the minimum memory requirement for the robot is $0.57$ $MegaBytes$, which is the minimum memory requirement for storing all the outputs of all the algorithms. Since the average communication time is larger than the execution time of most algorithms, it can be intuitively concluded that all algorithms should be allocated to the same node, and the most suitable node is a fog node due to the random delays. The average of the distances to the origin with our method is $1.041572$ with standard deviation of $0.01228939$ and with the method of \cite{li:2018} is $1.041576$ with standard deviation of $0.01229054$, for $50$ experiments. The code to our method is available on \url{https://github.com/SaeidZadeh/Static-Singlerobot.git}.

\section{Conclusion}
We have provided a model for users to decide which type of robot is better suited to perform their required tasks, and also how algorithms can be allocated to have the fastest overall performance in the context of cloud robotics. The suitability of a robot for a given task must be decided based on the memory capacity required, the processing time we can afford for a given task, and the budget available.

First, we minimized the maximum time that the robot takes between sending requests for any algorithm and receiving the response. Note that the maximum time is always for the last algorithm (virtual algorithm at the bottom of the graph of algorithms). Then we have minimized the total memory usage of the robot. Since the minimum time and minimum memory are not necessarily for the same allocation of algorithms, we identify lower bounds such that the allocation problems for memory and time intersect with respect to these bounds. If the intersection has more than one solution to the allocation problem, the decision to the allocation of algorithms can be made by a user depending on the importance of the memory usage or overall completion time factors.

Our method provides a solution for achieving optimal performance of the cloud robotic system with a single robot, and also allows comparison between the performances of multiple robots. Our method also works when we consider multiple fog and cloud nodes, and also when we consider a direct communication link between edge and cloud nodes. This is because the algebra of memory and the algebra of time are obtained regardless of the number of nodes in the architecture, and the communication time between two nodes in the architecture can be considered as the shortest communication time between the nodes. This work is also a first step towards solving the allocation problem for cloud robotic systems with multiple robots, which goes beyond the above shortcomings of the currently existing solutions.

\section*{Acknowledgment}

This work was supported by Opera\c{c}\~{a}o Centro - 01 - 0145 - FEDER - 000019 - C4 - Cloud Computing Competence Center, co-financed by the Programa Operacional Regional do Centro (CENTRO 2020), through the Sistema de Apoio \`{a} Investiga\c{c}\~{a}o Cientif\'{i}ca e Tecnol\'{o}gica - Programas Integrados de IC\&DT. This work was supported by NOVA LINCS (UIDB/04516/2020) with the financial support of FCT-Funda\c{c}\~{a}o para a Ci\^{e}ncia e a Tecnologia, through national funds. We also thank Andr\'{e} Correia for providing the real world data in Tables \ref{tab2p} and \ref{tab3p}.

\bibliographystyle{unsrt}
\bibliography{sample}

\section*{Biography}

\textbf{Saeid Alirezazadeh} received the Ph.D. degree from Centro de Matem\'{a}tica - Universidade do Porto, Porto, Portugal, in 2015. He was a Post-Doctoral Researcher with Instituto Superior de Agronomia, Lisbon, Portugal, from 2016 to 2019. He is currently a Post-Doctoral Researcher with C4-Cloud Computing Competence Center, Universidade da Beira Interior, Covilh\~{a}, Portugal. His research interests include algebraic graph theory, semigroups, automata, and tree languages, optimal scheduling of cloud robotics systems, and theoretical modeling of dynamical systems and stochastic processes.

\textbf{Lu\'{i}s A.~Alexandre} received his BSc (1994), MSc (1997) and PhD (2002) from the University of Porto, and an Habilitation on Computer Science and Engineering (2013) by UBI. His research interests are neural networks, computer vision and their applications, particularly to robotics. He has been PI of several research projects with total budgets over 2M\texteuro, involving both academia and industry, has co-chaired 2 international conferences, participated in more than 50 international conference program committees and has been an evaluator for the A3ES (portuguese agency for the Quality Assurance in Higher Education). He was also a member of the governing board of both the International Asssociation for Pattern Recognition and of the European Neural Network Society and the president of the Portuguese Association for Pattern Recognition.

\end{document}